\documentclass[10pt,journal,compsoc]{IEEEtran}
\ifCLASSOPTIONcompsoc
  \usepackage[nocompress]{cite}
\else
  \usepackage{cite}
\fi

\usepackage{amssymb,amsfonts}
\usepackage{times}
\usepackage{enumitem}   
\usepackage{bbm}
\usepackage{lipsum}
\usepackage{comment}
\usepackage{array}
\usepackage{mathtools} 
\usepackage{algorithm}
\usepackage{amsmath,amsthm,bm}
\usepackage{soul}
\usepackage{mdwmath}
\usepackage{mdwtab}
\usepackage{blindtext}
\usepackage{mathrsfs}
\usepackage{subcaption}
\usepackage{cite}
\usepackage{booktabs}
\usepackage{amsmath,amssymb,amsfonts}
\usepackage{algorithmic}
\usepackage{graphicx}
\usepackage{textcomp}
\usepackage{xcolor}
\usepackage{multirow}
\usepackage{graphicx}
\usepackage{lscape}
\newcommand*{\tran}{^{\mkern-1.5mu\mathsf{T}}}

\newcommand{\argmin}{\operatornamewithlimits{argmin}}

\usepackage[utf8]{inputenc}
\usepackage[english]{babel}
\newtheorem{theorem}{Theorem}

\newtheorem{lemma}{Lemma}
\newtheorem{definition}{Definition}
\usepackage{cite}
\usepackage{amsmath,amssymb,amsfonts}
\usepackage{algorithmic}
\usepackage{graphicx}
\usepackage{textcomp}
\usepackage{xcolor}
\usepackage{soul}

\begin{document}

\title{Differentially Private Federated Learning for Resource-Constrained Internet of Things}

\author{Rui~Hu,~\IEEEmembership{Student~Member,~IEEE,} Yuanxiong~Guo,~\IEEEmembership{Senior~Member,~IEEE,} ~E.~Paul~Ratazzi, and Yanmin~Gong,~\IEEEmembership{Member,~IEEE}
 \IEEEcompsocitemizethanks{\IEEEcompsocthanksitem R. Hu, Y. Guo and Y. Gong are with the University of Texas at San Antonio, San Antonio, TX, 78249. E-mail: \{rui.hu, yuanxiong.guo, yanmin.gong \}@utsa.edu.\protect
 \IEEEcompsocthanksitem E. P. Ratazzi is with Air Force Research Laboratory, Information Directorate, Rome, NY, 13441. E-mail: edward.ratazzi@us.af.mil.}
 }

\IEEEtitleabstractindextext{%
\begin{abstract}
With the proliferation of smart devices having built-in sensors, Internet connectivity, and programmable computation capability in the era of Internet of things (IoT), tremendous data is being generated at the network edge. Federated learning is capable of analyzing the large amount of data from a distributed set of smart devices without requiring them to upload their data to a central place. However, the commonly-used federated learning algorithm is based on stochastic gradient descent (SGD) and not suitable for resource-constrained IoT environments due to its high communication resource requirement. Moreover, the privacy of sensitive data on smart devices has become a key concern and needs to be protected rigorously. This paper proposes a novel federated learning framework called DP-PASGD for training a machine learning model efficiently from the data stored across resource-constrained smart devices in IoT while guaranteeing differential privacy. The optimal schematic design of DP-PASGD that maximizes the learning performance while satisfying the limits on resource cost and privacy loss is formulated as an optimization problem, and an approximate solution method based on the convergence analysis of DP-PASGD is developed to solve the optimization problem efficiently. Numerical results based on real-world datasets verify the effectiveness of the proposed DP-PASGD scheme.
\end{abstract}

}
\maketitle

\IEEEdisplaynontitleabstractindextext
\IEEEpeerreviewmaketitle

\IEEEraisesectionheading{\section{Introduction}\label{sec:intro}}
\IEEEPARstart{T}{he} proliferation of smart devices with built-in sensors, Internet connectivity, and programmable computation capability in the era of Internet of things (IoT) leads to tremendous data being generated at the network edge. This rich data, if collected, shared, and analyzed efficiently, can power a wide range of useful IoT applications, such as personal fitness tracking\cite{iotwearable}, traffic monitoring\cite{iottraffic}, and smart home security\cite{smarthome}, and renewable energy integration\cite{iotenergy}. Among the available methods in analyzing large amounts of data, machine learning is the state-of-the-art and enables learning statistical models from data for detection, classification, and prediction of future events. 

Traditional machine learning works mostly in a centralized way by first uploading all data to a central location (e.g., in the cloud) and then performing model training using some powerful servers. With the growth of the computation and storage capabilities of smart devices, constrained network bandwidth, and increasing privacy concerns associated with personal data, federated learning that stores data locally and trains models distributedly on each smart device is gaining popularity \cite{mcmahan2017communication}. Since raw data is kept locally on each device without being shared directly with the central server, federated learning can achieve higher efficiency and better privacy in comparison with centralized machine learning in IoT. 

As the core backbone of most state-of-the-art machine learning algorithms, (mini-batch) stochastic gradient descent (SGD) has been widely used in the federated learning setting \cite{dekel2012optimal}. In the distributed SGD, at the beginning of each communication round, a central server first sends the current model to each of the smart devices. Then each device calculates the gradient of the loss function based on the current model and a mini-batch of its local dataset and sends it back to the server. Next, the server aggregates these gradients and updates its global model, and the process repeats. Although computationally efficient, distributed SGD often requires a very large number of communication rounds between the smart devices and central server to reach a high accurate model \cite{mcmahan2017communication}, which is inefficient for resource-constrained IoT devices with expensive communication connections and limited battery sizes. 

To address the limitation of communication efficiency, distributed SGD algorithms with periodic averaging have been proposed in \cite{zhang2016parallel,mcmahan2017communication,wang2018cooperative,stich2018local}. The basic idea is to allow each device to perform multiple local updates to the model instead of just computing gradients and then periodically aggregate the local models. By performing more computation at smart devices between each communication round, those algorithms are shown to work well with fewer numbers of communication rounds \cite{zhang2016parallel,wang2018cooperative,stich2018local}. Since each local update consumes computation resource and each global aggregation consumes communication resource, the global aggregation period needs to be carefully chosen to balance the model accuracy and total resource consumption. 

While resource efficiency is a key concern for federated learning on IoT devices, protecting the privacy of participating users and their sensitive data is an equally important consideration. Note that the intermediate results (e.g., gradients or local models) exchanged during the federated learning process could leak private user information as demonstrated by recent attacks such as model inversion attacks \cite{fredrikson2015model} and membership attacks \cite{shokri2017membership}. Differentially private noise can be added into the intermediate results in the distributed learning algorithms to provide rigorous privacy guarantee, but it affects the model accuracy. When jointly considering the requirements of resource efficiency and privacy, there is a complicated relationship among the model accuracy, resource cost, and privacy for federated learning in IoT. 


There are several recent studies that focus on either reducing communication/computation resource usage \cite{wang2018edge,alistarh2017qsgd,tran2019federated,wangni2018gradient} or providing privacy guarantee \cite{mcmahan2018learning,guo2018practical,abadi2016deep,huang2012differentially}, but not both, in federated learning. When considering resource efficiency and privacy protection simultaneously, the impacts of the learning algorithm on the above two aspects needs to jointly analyzed and optimized to obtain the optimal schematic design, which is much more challenging than only considering a single aspect. Moreover, none of the previous studies in literature have rigorously investigated the resource allocation for federated learning over resource-constrained IoT, which needs to explicitly model the communication and computation resource and privacy constraints in IoT networks and has a large impact on the learning accuracy. 

Motivated by the above observations, in this paper we propose a novel federated learning framework called DP-PASGD that is easy to be implemented on resource-constrained IoT devices and guarantees differential privacy. The proposed framework integrates distributed SGD with periodic averaging to reduce resource cost and differentially private noise addition to preserve privacy. We then investigate the optimal configuration of DP-PASGD and propose an optimization framework to maximize the model accuracy while satisfying the limits on resource cost and privacy loss of IoT devices in DP-PASGD. Next, we develop an approximate solution approach to find the optimal configuration under the proposed framework efficiently. The problem is significant to resource-constrained IoT environments because the training of machine learning models is often resource-intensive, and a non-optimal learning scheme could quickly drain the resources of smart devices and violate users' privacy, discouraging them to participate. 


In summary, the main contributions of this paper are as follows.
\begin{itemize}
    \item We propose a novel federated learning framework called DP-PASGD for training a machine learning model both efficiently and privately from the data stored across resource-constrained smart devices in IoT.  

    
    \item We investigate the optimal schematic design of DP-PASGD in resource-constrained IoT environments and develop an optimization framework to balance the trade-offs among model accuracy, privacy, and resource cost. 
    
    \item We perform rigorous convergence analysis of DP-PASGD and leverage it to develop an approximate solution approach to find the optimal configuration of DP-PASGD efficiently.  
    
    
    \item We conduct extensive evaluations based on real-world datasets, verify the effectiveness of the proposed scheme, and observe the trade-offs among model accuracy, privacy, and resource cost empirically. 
\end{itemize}

The rest of the paper is organized as follows. Related work and background on privacy notations used in this paper are described in Section~\ref{sec:related} and Section~\ref{sec:pre}, respectively. Section~\ref{sec:sys-mod} introduces the system setting and the proposed DP-PASGD framework. Section~\ref{sec:pro-form} presents an optimization problem formulation for the optimal schematic design of DP-PASGD under resource-constrained IoT environments. The convergence property of DP-PASGD is rigorously analyzed in Section~\ref{sec:convergence}, and the solution approach to find the optimal configuration of DP-PASGD is developed in Section~\ref{sec:control}. Finally, Section~\ref{sec:eva} shows the evaluation results based on real-world datasets, and Section~\ref{sec:con} concludes the paper. 

\section{{Related works}} \label{sec:related}

Distributed machine learning based on SGD has been well studied in literature with both theoretical convergence analysis \cite{kushner2003stochastic,shamir2013stochastic,bottou2018optimization} and real-world experiments \cite{dean2012large}. However, traditional distributed SGD does not fit into the IoT setting wherein the communication cost is usually high and smart devices are often resource-constrained. Recent studies have started to reduce the resource usage, particularly communication resource, in SGD-based distributed learning \cite{alistarh2017qsgd,wang2018edge,tran2019federated}. Two most common approaches are (i) periodic model averaging that puts more computation on each device between each communication round \cite{wang2018edge,tran2019federated}; and (ii) gradient compression that quantizes and/or sparsifies gradients computed by each device \cite{alistarh2017qsgd,wangni2018gradient}. However, most of the proposed resource-efficient schemes ignore the privacy aspect and do not explicitly model resource constraints on smart devices. Our proposed scheme achieves both resource efficiency and rigorous privacy protection through integrating periodic model averaging and differential privacy. Moreover, we provide an optimization framework to balance the trade-offs among model accuracy, resource consumption, and privacy guarantee in the proposed scheme and solve it under practical resource and privacy constraints. Agarwal et al. \cite{agarwal2018cpsgd} also proposes a distributed SGD scheme that achieves both communication-efficiency and differential privacy, but it focuses on gradient compression and hence is orthogonal to our work. Moreover, it is not clear from their scheme how to maximize learning accuracy under differential privacy and communication-efficiency constraints.  

Differentially private distributed learning is also an active research area, and a wide range of differentially private algorithms have been proposed based on different distributed optimization algorithms (e.g., alternating direction method of multipliers (ADMM), gradient descent, and distributed consensus) and noise addition mechanisms (e.g., output perturbation, objective perturbation, and gradient perturbation) \cite{guo2018practical,abadi2016deep,huang2012differentially,jayaraman2018distributed,mcmahan2018learning,shokri2015privacy}. However, none of the known privacy-preserving schemes explicitly model or optimize the resource efficiency aspect, and if applied directly to IoT, could lead to sub-optimal performance. In comparison, we perform a rigorous convergence analysis of our proposed differentially private algorithm and use it to optimize the configuration of our algorithm under different IoT settings. 

\section{Background}\label{sec:pre}

Differential privacy (DP) is a cryptography-inspired rigorous notion of privacy and has become the de-facto standard for measuring privacy risk \cite{dwork2014algorithmic}. In this section, we briefly describe the basics of DP and their properties to be used in the rest of this paper.

\subsection{$(\epsilon, \delta)$-Differential Privacy}

$(\epsilon, \delta)$-DP is the classic DP notion with the following definition:

\begin{definition}[$(\epsilon,\delta)$-DP]\label{DP} 
A randomized algorithm $\mathcal{M}:\mathcal{D} \to \mathcal{R}$ with domain $\mathcal{D}$ and range $\mathcal{O}$ is $(\epsilon,\delta)$-differentially private if for any two adjacent datasets $D, D^{\prime} \subseteq \mathcal{D}$ that differ in at most one data sample and any subset of outputs $\mathcal{S} \subseteq \mathcal{O}$, it satisfies that:
\begin{equation}
\Pr[\mathcal{M}(D) \in \mathcal{S}] \leq e^{\epsilon} \Pr[\mathcal{M}(D^{\prime}) \in \mathcal{S}] + \delta.
\end{equation}
\end{definition}

The above definition reduces to $\epsilon$-DP when $\delta=0$. Here the parameter $\epsilon$ is also called the privacy budget. Given any function $f$ that maps a dataset $D\in\mathcal{D}$ into a vector ${o}\in \mathbb{R}^d$, we can achieve $(\epsilon,\delta)$-DP by adding Gaussian noise to each of the $d$ coordinates of the output vector ${o}$, where the noise is proportional to the sensitivity of $f$, given as $\Delta_2(f):= \|f(D) - f(D^\prime) \|_2$.
\subsection{Zero-Concentrated Differential Privacy}

Zero-concentrated differential privacy \cite{bun2016concentrated} (zCDP) is a relaxed version of $(\epsilon, \delta)$-DP. zCDP has a tight composition bound and is more suitable to analyze the end-to-end privacy loss of iterative algorithms. To define zCDP, we first define the privacy loss random variable. Given an output $ o \in \mathcal{R}$, the privacy loss random variable $Z$ of the mechanism $\mathcal{M}$ is defined as
\begin{equation}
Z:= \log\frac{\Pr[\mathcal{M}(D) = o]}{\Pr[\mathcal{M}(D^{\prime})=o]}.
\end{equation}
zCDP imposes a bound on the moment generating function of the privacy loss $Z$. Formally, a randomized mechanism $\mathcal{M} $ satisfies $\rho$-zCDP if for any two adjacent datasets $D, D^{\prime} \subseteq \mathcal{D}$, it holds that for all $\alpha\in(1,\infty)$,
\begin{equation}\label{zcdp-bound}
\mathbb{E}[e^{(\alpha-1)Z}] \leq e^{(\alpha-1)\rho}.
\end{equation}
Here, \eqref{zcdp-bound} requires the privacy loss $Z$ to be concentrated around zero, and hence it is unlikely to distinguish $D$ from $D^{\prime}$ given their outputs. zCDP has the following properties \cite{bun2016concentrated}:
\begin{lemma}\label{composition}
Suppose two mechanisms satisfy $\rho_1$-zCDP and $\rho_2$-zCDP, then their composition satisfies $\rho_1+\rho_2$-zCDP.
\end{lemma}
\begin{lemma}\label{rho-zcdp}
The Gaussian mechanism, which returns $f(\mathcal{A}) + \mathcal{N}(0,\sigma^{2})$, satisfies $\Delta_2(f)^2/(2\sigma^2)$-zCDP.
\end{lemma}
\begin{lemma}\label{zcdp-dp}
If $ \mathcal{M}$ is a mechanism that provides $\rho$-zCDP, then $ \mathcal{M}$ is $(\rho + 2\sqrt{\rho\log({1}/{\delta})}, \delta) $-DP for any $\delta > 0$.
\end{lemma}

\section{Federated Learning in IoT Networks}\label{sec:sys-mod}

\begin{figure}[ht]
\centering
\includegraphics[width=0.9\linewidth]{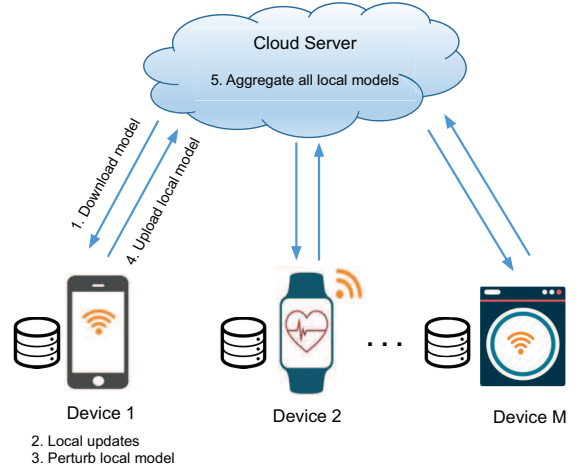}
\caption{IoT system architecture.}\label{fig:system_model}
\end{figure}

We consider an IoT network as depicted in Figure~\ref{fig:system_model}. In the system, a set of smart devices $\mathcal{M} := [1, \ldots, M]$ collect their own datasets and want to collaboratively learn a shared model over the entire data across all smart devices. Each device has some embedded computing capability to train a local model. A cloud server is responsible for coordinating the information exchange among the devices to learn the shared model. 


Assume each device $m \in \mathcal{M}$ has a training dataset $\mathcal{D}_m = \{(\mathbf{x}_n^{m},y_n^m), \forall n \in \mathcal{N}_m := [1, \ldots, N_m]\}$ with $\mathbf{x}_n^{m}$ and $y_n^{m}$ to be the feature vector and corresponding label of the $n$-th training sample, respectively. The goal is to learn a model with parameter vector $\boldsymbol{\theta}\in \mathbb{R}^d$ that can minimize the following empirical risk function:
\begin{equation}\label{problem}
\min_{\bm{\theta} \in \mathbb{R}^{d}} \quad  \mathcal{L}(\bm{\theta}):= \frac{1}{M}\sum_{m \in \mathcal{M}}\frac{1}{N_m} \sum_{ n\in \mathcal{N}_m} {l}(\bm{\theta}; \mathbf{x}_n^m,y_n^m). 
\end{equation}
Here ${l}(\cdot)$ measures the accuracy of the model on a data sample and is assumed to be a convex loss function with $G$-Lipschitz continuity and $L$-smoothness. Without loss of generality, we assume that each data sample lies in a unit ball which can be enforced through normalization. 

\subsection{Distributed SGD}

Distributed SGD\cite{dekel2012optimal,li2014scaling} is a popular way to minimize the objective function $\mathcal{L}(\bm{\theta})$ in a distributed setting. Using classic mini-batch SGD, updates to the model parameter vector $\bm{\theta}$ are performed as follows. Let $\mathcal{X}_m \subseteq \mathcal{D}_m$ be a mini-batch of device $m$'s dataset with size $X_m$. The update rule of distributed SGD at each iteration $k$ is
\begin{equation}\label{syn-sgd}
\bm{\theta}^{k} := \bm{\theta}^{k-1} - \eta \Big[ \frac{1}{M}\sum_{m \in \mathcal{M}} g(\bm{\theta}^{k-1}; \mathcal{X}_m) \Big],
\end{equation}
where each device $m$ computes a simple gradient $g(\bm{\theta}; \mathcal{X}_m):= (1/X_m) \sum_{n \in \mathcal{X}_m} \nabla l(\bm{\theta};\mathbf{x}_n^m,y_n^m)$ from a mini-batch of its local dataset, and the cloud server averages those gradients periodically and updates the model parameters with learning rate $\eta$. For notational simplicity, we will use $g(\bm{\theta})$ instead of $ g(\bm{\theta}; \mathcal{X}_m)$ in the rest of the paper. 

Although computationally efficient in each iteration (i.e., only a gradient is computed at each device), distributed SGD requires a large number of communication rounds between the smart devices and cloud server to achieve good model accuracy \cite{ioffe2015batch}, which is not feasible for resource-constrained IoT where communication resource is often the bottleneck. 

\subsection{Periodic Averaging SGD}


To address the limitations of distributed SGD, recent works propose the periodic averaging SGD (PASGD) framework to reduce the communication cost by allowing devices to perform more computation between each communication round. Specifically, each device performs $\tau$ local updates to the model parameters $\bm{\theta}$ instead of just computing gradient during each communication round, and then the resulting locally updated models (which are different due to variability in training data across devices) are averaged by the cloud server every $\tau$ iterations. The update rule of PASGD with global aggregation period $\tau$ at each iteration $k$ is:
\begin{subequations}\label{PASGD_update}
\begin{equation}\label{local_update_rule}
\bm{w}_m^{k} := \bm{\theta}_m^{k - 1} - \eta g(\bm{\theta}_m^{k - 1}),
\end{equation}
\begin{gather}\label{global_aggregation_rule}
\bm{\theta}_m^{k} :=
\begin{cases}
\frac{1}{M}\sum_{m \in \mathcal{M}} \bm{w}_m^{k}, & \text{ if } k \text{ mod } \tau = 0 \\
\bm{w}_m^{k}, &\text{ otherwise}
\end{cases}
\end{gather}
\end{subequations}
where $\bm{ \theta}_m^{k}$ denotes the learned local model parameters of device $m$ at iteration $k$ with $\bm{w}_m^{k}$ being the intermediate results, and $\eta$ denotes the learning rate. Extensive empirical results have validated the effectiveness of PASGD in improving the speed and scalability of distributed SGD when choosing an appropriate value of $\tau$ \cite{mcmahan2017communication,bonawitz2019towards}. 


\subsection{Differentially-Private PASGD}

Although communication-efficient, PASGD does not provide rigorous privacy guarantee for the participating devices and their sensitive information. Specifically, we consider the following attack model: the cloud server and smart devices are all ``honest-but-curious'', and the information exchanged through the network is secured during the transmission using standard security protocols. By observing the received local model of a victim device, it is possible for the cloud server or other devices to recover the private dataset of the victim device using reconstruction attack \cite{al2016reconstruction} or infer whether a sample is in the dataset of the victim with membership inference attack \cite{shokri2017membership}. Our design goal is to ensure that the cloud server or other devices cannot learn much additional information of the victim device's dataset from the exchanged messages during the execution of the learning scheme under any auxiliary information. 

We design our privacy-preserving PASGD under the framework of differential privacy \cite{dwork2014algorithmic}. A differentially private algorithm provides a strong guarantee that the presence of an individual record in the dataset will not significantly change the output of the algorithm. Specifically, we use the gradient perturbation where the gradients computed at each iteration are perturbed via adding Gaussian noise. The update rule of the differentially private PASGD (DP-PASGD) at iteration $k$ is as follows:
\begin{subequations}\label{DPPASGD_update}
\begin{equation}\label{local_update_rule_2}
\bm{w}_m^{k} := \bm{\theta}_m^{k - 1} - \eta \left(g(\bm{\theta}_m^{k - 1}) + \bm{b}_m^{k}\right),
\end{equation}
\begin{gather}\label{global_aggregation_rule_2}
\bm{\theta}_m^{k} :=
\begin{cases}
\frac{1}{M}\sum_{m \in \mathcal{M}} \bm{w}_m^{k}, & \text{ if } k \text{ mod } \tau = 0 \\
\bm{w}_m^{k}, &\text{ otherwise}
\end{cases}
\end{gather}
\end{subequations}
where $\mathbf{b}_m^k \sim \mathcal{N}(0, \mathbf{1}_d\sigma_m^2)$ represents the Gaussian noise with standard deviation $\sigma_m$, and other notations are the same as those in \eqref{local_update_rule}--\eqref{global_aggregation_rule}. The resulting protocol of DP-PASGD is depicted in Figure~\ref{fig:system_model}. Specifically, at the beginning of each communication round, each device first downloads a global model from the cloud server and then uses \eqref{local_update_rule_2} to update the local model and add Gaussian noise. Next, after $\tau$ local updates, the device send the updated noisy local model to the cloud server where all the received local models are aggregated to get the updated global model according to \eqref{global_aggregation_rule_2}. Finally, the process goes to the next communication round and repeats. 

\section{Optimal Design of DP-PASGD under Resource and Privacy Constraints}\label{sec:pro-form}

Although the proposed DP-PASGD have the potential to achieve high resource efficiency and differential privacy simultaneously, it is unclear how to configure DP-PASGD, such as the global aggregation period $\tau$, total number of iterations $K$, and noise magnitude $\sigma_m$, under the resource-constrained IoT setting where each device has certain limits on its resource cost and privacy budget for learning the model. Note that each local update consumes certain computation resource, and each global aggregation consumes certain communication resource. The privacy budget is consumed whenever the local dataset is queried to compute the gradient $g(\bm{\theta})$. It is obvious that the global aggregation period $\tau$ affects the total resource cost and final model accuracy after $K$ iterations. Moreover, the noise magnitude at each iteration $\sigma_m$ and the total number of iterations $K$ would determine the total privacy loss and affect the final model accuracy. Therefore, there are complex relationships among the DP-PASGD configuration variables (e.g., global aggregation period $\tau$, total number of iterations $K$, and noise magnitude $\sigma_m$), model accuracy, resource cost, and privacy guarantee. In the following, we rigorously model such relationships and propose an optimization framework to optimally select the DP-PASGD algorithmic parameters under resource and privacy constraints.  

\subsection{Resource Cost Model}

To analyze the effect of DP-PASGD configurations on the resource cost, we mainly focus on the communication and computation costs consumed by each device, which correspond to battery energy usage or running time in practice, during the learning process. Specifically, the communication cost is spent on uploading the local model and downloading the global model to and from the cloud server, respectively, and the computation cost comes from the local model update at each iteration. Assume the communication cost of each global aggregation step on a device is $c_1$, and the computation cost of each local update step on a device is $c_2$. Given the total number of iterations $K$ and global aggregation period $\tau$, the overall resource cost of a device is computed as
\begin{equation}\label{cost}
C = \frac{c_1 K}{\tau} + c_2 K,
\end{equation}
where we assume $K$ is an integer multiple of $\tau$. Here, larger $\tau$ implies less frequent global aggregation and smaller communication cost per iteration. We assume that $c_1$ and $c_2$ can be estimated beforehand in DP-PASGD. 

\subsection{Privacy Loss Model}

In order to analyze the impact of global aggregation period $\tau$ on privacy, we analyze the overall privacy loss of a device in DP-PASGD. For a device $m$, given any two neighboring datasets $\mathcal{X}_m$ and $\mathcal{X}_m^\prime$ of size $X_m$ that differ only in the $i$-th data sample, the sensitivity of the stochastic gradient computed at each iteration in PASGD can be computed as
\begin{multline*}
\|g(\bm{\theta}_m^{k}; \mathcal{X}_m) - g(\bm{\theta}_m^{k}; \mathcal{X}_m^\prime)\| \\
= \frac{1}{X_m} \|\nabla l(\bm{\theta}_m^{k}; \mathbf{x}_i^\prime,y_i^\prime) - \nabla l(\bm{\theta}_m^{k}; \mathbf{x}_i,y_i) \|. 
\end{multline*}
Since the loss function $l(\cdot)$ is $G$-Lipschitz continuous, the sensitivity of $g(\bm{\theta}_m^{k})$ can be estimated as $\Delta_2(g(\bm{\theta}_m^{k})) \leq 2G/X_m$. 
%

According to Lemma~\ref{rho-zcdp}, by adding $\mathbf{b}_m^k$ to each gradient $g(\bm{\theta}_m^{k})$, the DP-PASGD satisfies $(2G^2/ X_m^2\sigma_m^2)$-zCDP at each iteration. Using the composition result from Lemma~\ref{composition}, the DP-PASGD algorithm after $K$ iterations achieves $\rho_m$-zCDP for device $m$ where $\rho_m = 2KG^2/ X_m^2\sigma_m^2$. 
%
Then by Lemma \ref{zcdp-dp}, the DP-PASGD algorithm satisfies $(\epsilon, \delta)$-DP, where the overall privacy loss of device $m$ is
\begin{equation}
\label{eps_k}
\epsilon_m = \frac{2KG^2}{X_m^2\sigma_m^2} + \frac{2G}{X_m\sigma_m}\sqrt{2K\log\left(\frac{1}{\delta}\right)}.
\end{equation}
From the above equation, we can see a larger $K$ implies a larger privacy loss $\epsilon_m$, and therefore the choice of $\tau$ can implicitly influence the privacy loss by imposing an upper bound for $K$ according to the cost model \eqref{cost}. 

\subsection{Optimization Framework}

In practice, smart devices in IoT have limited resources and certain minimum privacy expectations. Therefore, a natural question is given some budgets on resource cost and privacy loss of each device, how to design the DP-PASGD scheme so that the empirical loss is minimized (or model accuracy is maximized). Let $\epsilon_{th}$ be the overall privacy budget and $C_{th}$ be the overall resource budget for each device. To efficiently utilize the limited resource and privacy budgets while maximizing the model accuracy, we formulate the following optimization problem to find the optimal design of DP-PASGD:
\begin{subequations}\label{control-model}
\begin{alignat}{3}\label{obj_control_model}
& \min_{\tau\in\mathbb{N}, K\in\mathbb{N}, \{\sigma_m\}_{m \in \mathcal{M}}} \quad \mathcal{L}(\bm{\theta}^*)\\ 
& \text{s.t.} \quad \frac{2KG^2}{X_m^2\sigma_m^2} + \frac{2G}{X_m\sigma_m}\sqrt{2K\log\left(\frac{1}{\delta}\right)} \leq \epsilon_{th}, \forall m \in \mathcal{M},\\
& \quad \quad \frac{c_1 K}{\tau} + c_2 K \leq C_{th},
\end{alignat}
\end{subequations}
where $ \bm{\theta}^*$ is the best model parameters obtained among $K$ iterations, i.e., $\bm{\theta}^* := \argmin_{1\leq k \leq K} \{\mathcal{L}(\bm{\theta}^k)\}$.

The above optimization problem is hard to solve due to the following two challenges. First, the objective function $\mathcal{L}(\bm{\theta}^*)$ in \eqref{obj_control_model} does not have an explicit form since it is impossible to derive the closed form of $\bm{\theta}^*$ after $K$ iterations due to the randomized nature of DP-PASGD, making the problem intractable. Second, the optimization variables $\tau$ and $K$ are both integer variables, making the problem highly non-convex. To solve the optimization problem~\eqref{control-model}, our basic idea is to first perform the convergence analysis of the DP-PASGD, and then use the derived convergence error bound of DP-PASGD after $K$ iterations as the approximation to the original objective function and reformulate the problem into a tractable one. After that, we relax the integer variables and develop a heuristic to solve the reformulated problem efficiently. 

\section{Convergence Analysis of DP-PASGD}\label{sec:convergence}

In this section, we analyze the convergence properties of DP-PASGD and find an approximation to the objective function $\min \mathcal{L}(\bm{\theta}^*)$ in the optimization problem~\eqref{control-model}. The convergence analysis is conducted under the following common assumptions, which are similar to previous works on the distributed SGD \cite{wang2018cooperative}:
\begin{enumerate}
    \item  Smoothness: $ \|\nabla \mathcal{L}(\mathbf{x}) -\nabla \mathcal{L}(\mathbf{y}) \| \leq L\|\mathbf{x}-\mathbf{y}\|$;

    \item Strongly convex: $\frac{1}{2}\|\nabla \mathcal{L}(\mathbf{x})\|^2 \geq \lambda (\mathcal{L}(\mathbf{x}) - \mathcal{L}^*)$;


    \item Unbiased gradients: $ \mathbb{E}_{\mathcal{X}_m | \mathbf{x}}[g(\mathbf{x})] = \nabla \mathcal{L}(\mathbf{x})$;

    \item Bounded variances: $\mathbb{E}_{\mathcal{X}_m | \mathbf{x}}[ \| g(\mathbf{x) -  \nabla \mathcal{L}(\mathbf{x})}\|^2] \leq \xi^2$, where $\xi^2$ is inversely proportional to the mini-batch size.
\end{enumerate}
In the error-convergence analysis of DP-PASGD, we use the expected optimality gap over the distribution of the whole dataset of all devices as the convergence criteria, i.e., the algorithm achieves a $\gamma$-suboptimal solution if:
\begin{equation}
\mathbb{E}\Big[ \mathcal{L}(\bm{\theta}^*) - \mathcal{L}^*\Big] \leq \gamma,
\end{equation}
where $\gamma$ is an arbitrarily small value and $\mathcal{L}^*$ is the minimum loss. 
Specifically, we have the following main convergence results: 

\begin{theorem}[Convergence Error Bound of DP-PASGD] \label{convergence-1}
For the DP-PASGD algorithm, suppose the total number of iterations $K$ can be divided by the global aggregation period $\tau$. Under Assumptions $1)-4)$, if the learning rate satisfies $\eta L + \eta^2L^2\tau(\tau-1) \leq 1$, and all devices are initialized at the same point $\bm{\theta}^0 \in \mathbb{R}^d$. Then after $K$ iterations, the expected optimality gap is bounded as
\begin{equation}\label{conv_bound}
\mathbb{E}\Big[\mathcal{L}(\bm{\theta}^*) - \mathcal{L}^*\Big] \leq  (1-{\eta \lambda})^K \Big(\frac{\alpha - B}{K}\Big)+ B,
\end{equation}
where $\alpha := \mathcal{L}({\bm{\theta}}^{0}) - \mathcal{L}^*$ and 
\begin{equation}
B := \frac{\eta L + \eta^2L^2(\tau-1)M}{2\lambda M} \left(\xi^2 + \frac{d}{M}\sum_{m \in \mathcal{M}} \sigma_m^2\right).
\end{equation}
Here, $\xi^2$ is the variance bound of mini-batch stochastic gradients, $\sigma_m^2$ is the variance of Gaussian noise added for device $m$, $L$ is the Lipschitz constant of the gradient, $\lambda$ is the constant of strongly convexity and $M$ is the number of devices.
\end{theorem}

\begin{IEEEproof}
First of all, we present a general update rule that combines all the updating features in \eqref{local_update_rule_2}--\eqref{global_aggregation_rule_2}. Define matrices $\bm{\Theta}^k, \mathbf{G}^k, \mathbf{B}^k \in \mathbb{R}^{d\times M}$ that concatenate all local models, gradients and noises at iteration $k$:
\begin{equation*}
\begin{split}
   & \bm{\Theta}^k := [\bm{\theta}_1^{k}, \bm{\theta}_2^{k},  \dots, \bm{\theta}_M^{k} ],\\
   & \mathbf{G}^k := [ g(\bm{\theta}_1^{k}),g(\bm{\theta}_2^{k}), \dots, g(\bm{\theta}_M^{k})],\\
   & { \mathbf{B}^k := [\mathbf{b}_1^k,\mathbf{b}_2^k, \dots, \mathbf{b}_M^k]}.
\end{split}
\end{equation*}
Besides, define matrix $\mathbf{J} := \mathbf{1}\mathbf{1}^{\tran}/(\mathbf{1}^{\tran}\mathbf{1})$. Unless otherwise stated, $\mathbf{1}$ is an all-one column vector of size $M$, and the matrix $ \mathbf{J}$ and identity matrix $\mathbf{I}$ are of size $M\times M$. 

To capture periodic averaging, we define $\mathbf{J}^k$ as
\begin{gather*}
\mathbf{J}^k:= 
\begin{cases}
{ \mathbf{J}}, & \text{ if }k \text{ mod } \tau =0\\
\mathbf{I}_{M\times M}, & \text{ otherwise}
\end{cases}
\end{gather*}
and then the general update rule of DP-PASGD can be represented as follows:
\begin{equation}\label{averaged_model}
\bm{\Theta}^{k} =[\bm{\Theta}^{k-1} - \eta (\mathbf{G}^{k-1} + { \mathbf{B}^k)}] \mathbf{J}^k.
\end{equation}
Multiplying $(1/M)\textbf{1}$ on both sides of \eqref{averaged_model}, we have
\begin{equation*}
\frac{\bm{\Theta}^{k}\mathbf{1}_M}{M}=\frac{\bm{\Theta}^{k-1}\mathbf{1}_M}{M} - \eta\big[ \frac{ \mathbf{G}^{k-1}\mathbf{1}_M}{M} + { \frac{ \mathbf{B}^k\mathbf{1}_M}{M}}\big].
\end{equation*}
Define the averaged model at iteration $k-1$ as
\begin{equation}
\bar{\bm{\theta}}^{k-1} := \frac{\bm{\Theta}^{k-1}\mathbf{1}}{M} = \frac{1}{M} \sum_{m \in \mathcal{M}} \bm{\theta}_m^{k-1},
\end{equation}
and substituting the above equation into \eqref{averaged_model}, one yields
\begin{equation}
\bar{\bm{\theta}}^{k} = \bar{\bm{\theta}}^{k-1} - \eta \Big[\frac{1}{M}\sum_{m \in \mathcal{M}} \big( g(\bm{\theta}_m^{k-1}) +{  \mathbf{b}_m^k}\big)\Big].
\end{equation}

Let $\widetilde{g}(\bm{\theta}_m^{k-1})  := g(\bm{\theta}_m^{k-1}) + \mathbf{b}_m^k$, $\mathcal{G}^{k-1} := (1/M)\sum_{m \in \mathcal{M}} g(\bm{\theta}_m^{k-1})$, $ \widetilde{\mathcal{G}}^{k-1} := (1/M)\sum_{m \in \mathcal{M}}\widetilde{g}(\bm{\theta}_m^{k-1})$, $\mathcal{B}^k := (1/M)\sum_{m \in \mathcal{M}}\mathbf{b}_m^k$ and ${\mathcal{H}}^{k-1} := (1/M)\sum_{m \in \mathcal{M}} \nabla \mathcal{L}(\bm{\theta}_m^{k-1})$. Moreover, let $\nabla\mathcal{L}(\bm{\Theta}^{k-1}) := [\nabla\mathcal{L}(\bm{\theta}_1^{k-1}), \dots, \nabla\mathcal{L}(\bm{\theta}_M^{k-1})]$ and then we have $ \|\nabla\mathcal{L}(\bm{\Theta}^{k-1})\|_{F}^2 =  \sum_{m \in \mathcal{M}} \| \nabla\mathcal{L}(\bm{\theta}_m^{k-1})\|^2$ where $\|\cdot\|$ and $ \|\cdot\|_F$ denote the $L_2$-norm and Frobenius matrix norm, respectively. 

According to assumption 1), we have
\begin{align*}
    &\mathbb{E}\big[\mathcal{L}(\bar{\bm{\theta}}^{k}) -\mathcal{L}(\bar{\bm{\theta}}^{k-1})\big] \\
    &= -\eta \mathbb{E}\big[\langle \nabla\mathcal{L}(\bar{\bm{\theta}}^{k-1}), {  \widetilde{\mathcal{G}}^{k-1} }\rangle\big]  + \frac{\eta^2 L}{2} \mathbb{E}\big[\|{  \widetilde{\mathcal{G}}^{k-1} }\|^2 \big]\\
    & = -\eta \Big(\frac{1}{M}\sum_{m=1}^{M} \langle \nabla\mathcal{L}(\bar{\bm{\theta}}^{k-1}), \mathbb{E}\big[ \widetilde{g}(\bm{\theta}_m^{k-1}) \big]\rangle \Big)  + \frac{\eta^2 L}{2} \mathbb{E}\big[\|{  \widetilde{\mathcal{G}}^{k-1} }\|^2]\\
    & = -\frac{\eta}{2}\|\mathcal{L}(\bar{\bm{\theta}}^{k-1})\|^2 - \frac{\eta}{2M}\sum_{m=1}^{M} \|\nabla \mathcal{L}(\bm{\theta}_m^{k-1})\|^2 \\
    &\ \ +  \frac{\eta L^2}{2M}\sum_{m=1}^{M} \|\bar{\bm{\theta}}^{k-1} - \bm{\theta}_m^{k-1}\|^2 + \frac{\eta^2 L}{2} \mathbb{E}\big[\|{  \widetilde{\mathcal{G}}^{k-1} }\|^2].
\end{align*}
According to assumption 2), we obtain that 
\begin{align*}
    &\mathbb{E}\big[\mathcal{L}(\bar{\bm{\theta}}^{k}) -\mathcal{L}(\bar{\bm{\theta}}^{k-1})\big] \leq  \frac{\eta^2 L}{2} \mathbb{E}\big[\|{  \widetilde{\mathcal{G}}^{k-1} }\|^2] -\frac{\eta\|\nabla \mathcal{L}(\bm{\Theta}^{k-1})\|_{F}^2}{2M} \\
   &\qquad +  \frac{\eta L^2}{2M}\sum_{m=1}^{M} \|\bar{\bm{\theta}}^{k-1} - \bm{\theta}_m^{k-1}\|^2-{\eta \lambda}\mathbb{E}\big[\mathcal{L}(\bar{\bm{\theta}}^{k-1}) - \mathcal{L}^*\big].
\end{align*}
To simplify the notation, we take the last three terms of above inequality as $T_{k-1}$.
Taking the total expectation and averaging over $K$ iterations, one can obtain
\begin{multline}
    \label{stepconvex}
   \mathbb{E}\Big[\frac{1}{K} \sum_{k=1}^{K} \mathcal{L}(\bar{\bm{\theta}}^{k})  -\mathcal{L}^* - \frac{1}{K} \sum_{k=1}^{K}\frac{T_{k-1}}{\eta\lambda}\Big]  \leq \\
    (1-{\eta \lambda}) \mathbb{E}\Big[\frac{1}{K} \sum_{k=1}^{K}\mathcal{L}(\bar{\bm{\theta}}^{k-1}) - \mathcal{L}^* - \frac{1}{K} \sum_{k=1}^{K}\frac{T_{k-1}}{\eta\lambda} \Big].
\end{multline}
Assume $(1/K)\sum_{k=1}^{K}(T_{k-1}/\eta\lambda)$ is bounded by a constant $ B$. Then, by applying \eqref{stepconvex} repeatedly through iteration $K$, one yields
\begin{equation}
\label{conv}
     \mathbb{E}\Big[\frac{1}{K} \sum_{k=1}^{K} \mathcal{L}(\bar{\bm{\theta}}^{k})  -\mathcal{L}^* \Big]
    \leq \frac{1}{K}(1-{\eta \lambda})^K \Big(\alpha - B \Big) + B,
\end{equation}
where $\alpha := \mathcal{L}(\bar{\bm{\theta}}^{0}) - \mathcal{L}^*$.
 
Next, our goal is to find the upper bound $B$. Given that
\begin{align}
    \nonumber&\frac{1}{K} \sum_{k=1}^{K}\frac{T_{k-1}}{\eta\lambda} =  
     \frac{L^2}{2\lambda KM}\sum_{k=1}^K \sum_{m=1}^{M}\mathbb{E}\big[ \|\bar{\bm{\theta}}^{k-1} - \bm{\theta}_m^{k-1}\|^2 \big]\\
   &\ \ + \frac{-1}{2\lambda KM}\sum_{k=1}^{K}\|\nabla \mathcal{L}(\bm{\Theta}^{k-1})\|_{F}^2 +\frac{\eta L}{2\lambda K} 
    \sum_{k=1}^K\mathbb{E}\big[\|{ \widetilde{\mathcal{G}}^{k-1} }\|^2] ,\label{B}
\end{align}
we first bound the squared norm of perturbed stochastic gradient $\mathbb{E}\big[\|{ \widetilde{\mathcal{G}}^{k-1} }\|^2] $. Under Assumption 3) and 4), we have
\begin{align*}
    \mathbb{E}\big[\| { \widetilde{\mathcal{G}}^{k-1}} \|^2 \big] &= \mathbb{E}\big[\| { \widetilde{\mathcal{G}}^{k-1} }- \mathcal{H}^{k-1}\|^2 \big] + \|\mathbb{E}\big[{ \widetilde{\mathcal{G}}^{k-1}}]\|^2 \\
    & = \mathbb{E}\big[\| {\mathcal{G}}^{k-1} - \mathcal{H}^{k-1} + \mathcal{B}^k\|^2 \big] + \|\mathcal{H}^{k-1}]\|^2\\
    & = \mathbb{E}\big[\| {\mathcal{G}}^{k-1} - \mathcal{H}^{k-1} \|^2 \big] + { \mathbb{E}\big[\| \mathcal{B}^k\|^2 \big]} + \|\mathcal{H}^{k-1}]\|^2\\
    & = \frac{\xi^2 + d\overline{\sigma}^2}{M} + \frac{\|\nabla\mathcal{L}(\bm{\Theta}^{k-1})\|_{F}^2}{M},
\end{align*}
where $\overline{\sigma}^2 := 1/M\sum_{m=1}^{M} \sigma_m^2$ represents the average variance of Gaussian noises.

Then, we derive the upper bound of the network error term $\sum_{m=1}^{M}\mathbb{E}\big[ \|\bar{\bm{\theta}}^{k-1}- \bm{\theta}_m^{k-1}\|^2 \big]$. In order to facilitate the analysis, we first introduce some useful notations. Let ${ \widetilde{\mathbf{G}}^s:= [\widetilde{g}(\bm{\theta}_1^{s}),\dots, \widetilde{g}(\bm{\theta}_M^{s})]}$. Assume $ k-1=j\tau+i$, let  $ \mathbf{Y}_r := \sum_{s=\tau r+1}^{(r+1)\tau} \widetilde{\mathbf{G}}^s$ when $0 \geq r < j $ and $ \mathbf{Y}_r := \sum_{s=r\tau+1}^{r\tau + i -1} \widetilde{\mathbf{G}}^s$ when $r=j$, $ \mathbf{Q}_r := \sum_{s=r\tau +1}^{(r+1)\tau} \nabla \mathcal{L}(\bm{\Theta}^s)$ when $0 \geq r < j $ and $ \mathbf{Q}_r := \sum_{s=r\tau +1}^{r\tau + i -1} \nabla \mathcal{L}(\bm{\Theta}^s)$ when $ r=j$. Then, according to Equation (88) in \cite{wang2018adaptive}, we have 
\begin{align*}
    &\sum_{m=1}^{M} \mathbb{E}\|\bar{\bm{\theta}}^{k-1} - \bm{\theta}_m^{k-1} \|^2 \leq 2\eta^2 \underbrace{\mathbb{E}\Big[ \| \sum_{r=0}^{j} \mathbf{Q}_r(\mathbf{J}^{(j-r)} - \mathbf{J})\|_F^2 \Big]}_{T_2}\\
    &\qquad\qquad\qquad\qquad+ 2\eta^2 \underbrace{\mathbb{E}\Big[ \| \sum_{r=0}^{j} (\mathbf{Y}_r - \mathbf{Q}_r )(\mathbf{J}^{(j-r)} - \mathbf{J})\|_F^2 \Big]}_{T_1}.
\end{align*}
Based on Lemma 7 and Lemma 9 in \cite{wang2018cooperative}, we have
\begin{align*}
    T_1 &\leq \sum_{r=0}^{j} \mathbb{E}\big[ \|\mathbf{Y}_r - \mathbf{Q}_r \|_F^2\big]\|\mathbf{J}^{(j-r)} - \mathbf{J}\|_{op}^2 \\
    &= { \mathbb{E}\big[ \|\mathbf{Y}_j - \mathbf{Q}_j\|_F^2\big]} \\
    &=  \mathbb{E}\big[  \sum_{s = j\tau +1}^{j\tau + i -1} \sum_{m=1}^{M}  \|{ \widetilde{ g}(\bm{\theta}_{m}^s)} -  \nabla \mathcal{L}(\bm{\theta}_m^s)\|_F^2\big] \\
    & \leq (i-1) M ( \xi^2 + d \overline{\sigma}^2),
\end{align*}
where $\|\cdot\|_{op}$ is the matrix operation norm. Similar to the proof of $T_1$, we have
\begin{align*}
    T_2 &\leq \sum_{r=0}^{j} \mathbb{E}\big[\|\mathbf{Q}^r\|_F^2 \|\mathbf{J}^{(j-r)-\mathbf{J}}\|_{op}^2\big] = \mathbb{E}\big[\|\mathbf{Q}^j\|_F^2\big] \\
    &\leq (i-1) \sum_{s=j\tau+1}^{j\tau+i-1} \|\nabla\mathcal{L}(\bm{\Theta}^{s})\|_F^2.
\end{align*}
Accordingly, the network error over $K$ iterations is bounded by
\begin{align*}
    &\sum_{k=1}^K \sum_{m=1}^{M}\mathbb{E}\Big[ \|\bar{\bm{\theta}}^{k-1} - \bm{\theta}_m^{k-1}\|^2\Big]
    \leq
    2\eta^2\sum_{j=0}^{K/\tau -1} \sum_{i=1}^{\tau} (i-1) M ( \xi^2 \\
    & \qquad\qquad+ d \overline{\sigma}^2 ) +
    2\eta^2 \sum_{j=0}^{K/\tau -1} \sum_{i=1}^{\tau} \big( i-1\big) \sum_{s=j\tau+1}^{j\tau+i-1} \|\nabla\mathcal{L}(\bm{\Theta}^{s})\|_F^2 \\
    &\leq \eta^2\tau(\tau-1)\sum_{k=1}^K \|\nabla\mathcal{L}(\bm{\Theta}^{k-1})\|_F^2 + \eta^2{KM}(\tau-1)(\xi^2 + d \overline{\sigma}^2 ).
\end{align*}

Substituting the expression of network error and squared norm of gradient back to \eqref{B}, we obtain
\begin{align*}
    &\frac{1}{K} \sum_{k=1}^{K}\frac{T_{k-1}}{\eta\lambda} \leq   \frac{\eta L + \eta^2L^2(\tau-1)M}{2\lambda M} (\xi^2 + d\overline{\sigma}^2) \\
    &\ \  + \frac{1}{2\lambda KM}\Big[ \eta L + \eta^2 L^2 \tau (\tau-1) -1 \Big] \sum_{k=1}^K \|\nabla\mathcal{L}(\bm{\Theta}^{k-1})\|_F^2.
\end{align*}
If the learning rate satisfies $ \eta L + \eta^2 L^2 \tau (\tau-1)  \leq 1$, we have
\begin{equation}
\label{b-1}
    B = \frac{\eta L + \eta^2L^2(\tau-1)M}{2\lambda M} (\xi^2 + d\overline{\sigma}^2).
\end{equation}
Finally, since $ \mathbb{E}\big[\mathcal{L}(\bm{\theta}^{*}) - \mathcal{L}^*\big] \leq \mathbb{E}\big[1/K \sum_{k=1}^{K} \mathcal{L}(\bm{\theta}^{k})  -\mathcal{L}^* \big]$, Theorem \ref{convergence-1} follows by substituting \eqref{b-1} into \eqref{conv}.
\end{IEEEproof}

We can observe from Theorem~\ref{convergence-1} that the convergence error bound is dependent on the global aggregation period $\tau$, total number of iterations $K$, and noise magnitude $\sigma_m$. In particular, when $\tau=1$ and $\sigma_m = 0, \forall m\in \mathcal{M}$, the bound in \eqref{conv_bound} reduces to the bound of distributed SGD. Assume the learning rate satisfies $0 < 1-\eta\lambda < 1$. When $\tau$ increases, it enlarges the variance of local stochastic gradients which implies larger divergence among local models, and the bound will monotonically increase along with $\tau$. Similarly, the bound will increase proportional to the variance of noises $\sigma_m^2$, because the added Gaussian noises enlarge the divergence among local models at each iteration. In addition, it is straightforward to see that the decrease of the total iteration number $K$ will increase the bound as well.

\section{Approximate Solution  Approach}\label{sec:control} 

In this section, we use the upper bound derived in Theorem~\ref{convergence-1} to reformulate the original problem \eqref{control-model} into a tractable one and present an efficient algorithm to solve the reformulated problem.  

Given a loss function $\mathcal{L}(\bm{\theta})$, the minimum loss $\mathcal{L}^*$ is a constant. Thus, we can use the upper bound of $\mathbb{E}[\mathcal{L}(\bm{\theta}^*) - \mathcal{L}^*]$ as an approximation of $\mathcal{L}(\bm{\theta}^*)$ in Problem~\eqref{control-model} and obtain the following reformulated problem:
\begin{subequations}
\label{control-model-2}
\begin{alignat}{4}
&\min_{\tau \in \mathbb{N}, K\in\mathbb{N},  \{\sigma_m\}_{ m \in \mathcal{M}}} F := (1-{\eta \lambda})^K \Big({}\frac{\alpha - B}{K}\Big)+ B, \\ 
&\text{s.t. }  B = \frac{\eta L + \eta^2L^2(\tau-1)M}{2\lambda M} \left(\xi^2 + \frac{d}{M}\sum_{m=1}^{M}\sigma_m^2\right), \label{con:19_b}\\
&\quad \frac{2KG^2}{X_m^2\sigma_m^2} + \frac{2G}{X_m\sigma_m}\sqrt{2K\log\left(\frac{1}{\delta}\right)} \leq \epsilon_{th}, \forall m \in\mathcal{M}\\
& \quad \frac{c_1 K}{\tau} + c_2 K \leq C_{th},\\
&\quad \eta L + \eta^2L^2\tau(\tau-1) \leq 1. \label{con:learn_rate}
\end{alignat}
\end{subequations}

To solve the above mixed-integer non-linear problem, we first relax $\tau$ and $K$ to be real variables. Assume the learning rate $\eta$ is chosen to be small enough so that the constraint~\eqref{con:learn_rate} is satisfied and $1-\eta\lambda \in(0,1)$. Then by taking the gradient of the objective $F$ with respect to $\tau$ and using \eqref{con:19_b}, we have 
\[
\frac{\partial F}{\partial \tau} = \left(1-\frac{(1-\eta\lambda)^K}{K}\right)\frac{\eta^2L^2}{2\lambda}\left(\xi^2 + \frac{d}{M}\sum_{m=1}^{M}\sigma_m^2\right),
\]
which is positive. Thus, the objective monotonically increases with $\tau$, and the optimal $\tau^*$ is 
\begin{equation}\label{opt_tau}
\tau^* = \frac{c_1K}{C_{th}-c_2K}.
\end{equation}
Similarly, by taking the gradient of the objective $F$ with respect to $\sigma_m^2$, we obtain 
\[
\frac{\partial F}{\partial \sigma_m^2} = \Big(1-\frac{(1-\eta\lambda)^K}{K}\Big)\frac{\eta L + \eta^2 L^2 (\tau -1) d}{2\lambda M},
\]
which is always positive. Therefore, the objective monotonically increases with $\sigma_m^2$ and hence the optimal $\sigma_m^*$ satisfies
\[
\frac{2KG^2}{X_m^2(\sigma_m^*)^2} + \frac{2G}{X_m\sigma_m^*}\sqrt{2K\log\left(\frac{1}{\delta}\right)} = \epsilon_{th}.
\]
By solving the above equation, we have
\begin{equation}\label{opt_sigma}
(\sigma_m^*)^2 = \frac{2KG^2}{X_m^2 (\epsilon_{th} + 2\log(\frac{1}{\delta}) + 2\sqrt{(\log(\frac{1}{\delta}))^2 + \epsilon_{th} \log (\frac{1}{\delta})})}.
\end{equation}


After substituting the optimal value of $\tau^*$ and $(\sigma_m^*)^2$ into the objective function in problem \eqref{control-model-2} and rearranging the terms, we obtain the following relaxed form of problem \eqref{control-model-2} with variable $K$:
\begin{multline}
\label{control-model-3}
\min_{K \in \mathbb{R}}  \frac{\alpha (1-{\eta \lambda})^K}{K} + \Big(1-\frac{(1-{\eta \lambda})^K}{K}\Big) \Big(\frac{\eta L}{2\lambda M} \\
\quad + \frac{\eta^2 L^2(\frac{c_1K}{C_{th}-c_2K}-1)}{2\lambda} \Big)\Big(\xi^2 + \frac{2KdG^2}{MZ}\sum_{m=1}^{M} \frac{1}{X_m^2}\Big),
\end{multline}
where 
\begin{equation}
Z := \epsilon_{th} + 2\log(\frac{1}{\delta}) + 2\sqrt{(\log(\frac{1}{\delta}))^2 + \epsilon_{th} \log (\frac{1}{\delta})}
\end{equation}
is a constant. The problem~\eqref{control-model-3} is easily solvable by using the standard gradient descent algorithm. After obtaining the optimal solution $K^*$, we can calculate $\tau^*$ and $\sigma_m^*$ correspondingly based on the above analysis. Since $\tau$ and $K$ are integers in the original problem, we adopt a simple heuristic by rounding $K^*$ and $\tau^*$ to the nearest integers as the final solution. We will show in the numerical evaluation that such a heuristic has minimal impact on the solution accuracy.  

\section{Numerical Evaluation}\label{sec:eva}
In this section, we evaluate the performance of our proposed scheme DP-PASGD. We first describe our experimental setup and then show the efficiency of DP-PASGD in resource-constrained settings by comparing it with a baseline. Next, we show that our approximate solution method for the optimal design of DP-PASGD is effective by comparing our derived solution with the global optimal one obtained by the brute-force method. Finally, we show the trade-offs among resource cost, privacy, and model accuracy in DP-PASGD. 

\subsection{Experimental Setup}\label{subsec:exp_setup}
\textbf{Datasets and Learning Tasks.} We explore two real-world datasets using both logistic regression and SVM models in our experiments. 
The first dataset, \emph{Adult}\cite{blake1998uci}, contains 32,561 samples with 14 numerical and categorical features with each sample corresponding to a person. The task is to predict if the person's income exceeds $\$50,000$ based on the 14 attributes, namely, \textit{age, workclass, fnlwgt, education, education-num, marital-status, occupation, relationship, race, sex, capital-gain, capital-loss, hours-per-week, and native-country}.
To simulate a non-i.i.d. data distribution setting based on the Adult dataset, we first split it into 16 domains based on the \textit{education} attribute (i.e., Bachelors, Some-college, 11th, HS-grad, Prof-school, Assoc-acdm, Assoc-voc, 9th, 7th-8th, 12th, Masters, 1st-4th, 10th, Doctorate, 5th-6th, or Preschool) and then assign the data samples corresponding to each domain to a different device (16 devices in total). In this case, the average and standard deviation of the number of samples per device are 2,035 and 4,367, respectively. This non-i.i.d. setting is named as \textbf{Adult-1}. We also create an i.i.d. data distribution setting named as \textbf{Adult-2} by evenly assigning the original Adult data to 16 devices such that each device has 2,035 samples. 

The second dataset, \emph{Vehicle}\cite{duarte2004vehicle}, is collected from a distributed sensor network and contains acoustic, seismic, and infrared sensing data collected from 23 sensors. Each data sample consists of 100 features and a binary label (i.e., AAV-type or DW-type that represents the type of vehicle). To simulate a non-i.i.d. data distribution setting, we model each sensor as an individual device and assign its collected data to that specific device. The average and standard deviation of the number of samples per device are 1,899 and 349, respectively. The non-i.i.d. setting is called \textbf{Vehicle-1}. Similarly, we create the i.i.d. setting of the Vehicle dataset named \textbf{Vehicle-2} by evenly assigning all data in the Vehicle dataset to 23 devices so that each device has 1,899 samples. 

For the Adult-1 and Adult-2 cases, we train a logistic regression classifier on the 16 devices with just the categorical features to predict if the person's income exceeds $\$50,000$ or not and use the softmax cross-entropy as the loss function. For the Vehicle-1 and Vehicle-2 cases, we train a linear SVM on all of the 23 devices to predict whether a vehicle is AAV-type or DW-type and use the hinge loss as the loss function. 

\textbf{Baseline.} We select the state-of-the-art differentially private learning scheme named DP-SGD \cite{abadi2016deep} as a strong baseline to evaluate the efficiency of our proposed scheme. In DP-SGD, only one step of stochastic gradient descent is conducted to update the local model on each device during each aggregation period, and Gaussian noise is added to each update before sending it out for preserving the privacy of each device. 


\textbf{Hyperparameters.} We take $80\%$ of the data on each device for training, $10\%$ for testing and $10\%$ for validation. We tune the hyperparameters on the validation set and report the average accuracy on the testing sets of all devices. 
We take the initial loss as the value of the initial loss gap $\alpha$ and estimate the value of Lipschitz constant of gradient $L$, strongly convexity constant $\lambda$, initial loss gap $\alpha$, and variance bound of stochastic gradient $\xi^2$ beforehand. For all cases, we set communication cost per round $c_1 = 100$ and computation cost per iteration $c_2 = 1$ based on the typical setting of federated learning \cite{konevcny2016federated} and privacy failure probability $\delta = 10^{-4}$ by default. Note that due to the randomized nature of differentially private mechanisms, we repeat all the experiments for 5 times and report the average results. 

\subsection{Resource Efficiency of DP-PASGD}\label{subsec:efficiency}
\begin{figure}[ht]
\centering
\includegraphics[width=0.9\linewidth]{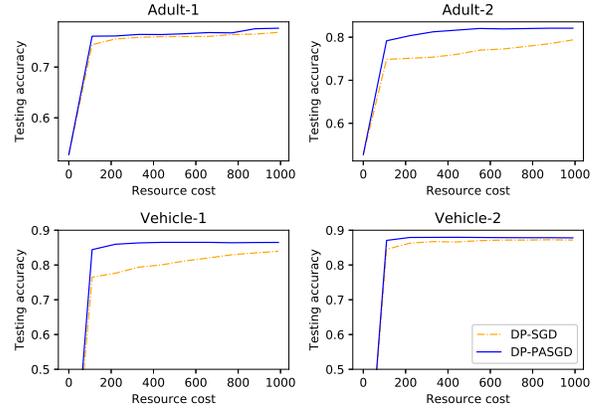}
\caption{Resource efficiencies of DP-PASGD ($\tau=10$) and DP-SGD when the maximum resource cost $C=1000$ and privacy loss $\epsilon=10$.}\label{fig:efficiency}
\end{figure}

In this subsection, we compare our proposed DP-PASGD with the baseline DP-SGD to show its resource efficiency. Specifically, we run the learning process of each scheme until reaching the maximum resource cost $C=1000$ and privacy loss $\epsilon=10$. In each global aggregation period, DP-SGD will run one local update while DP-PASGD will run 10 local updates (i.e., $\tau = 10$) on each device. 
The testing accuracies of both methods with respect to the resource cost are shown in Figure~\ref{fig:efficiency}.
%
We can observe that for all of the data distribution cases, DP-PASGD always achieves higher accuracy than DP-SGD. Hence, DP-PASGD with $\tau = 10$ achieves higher resource efficiency than DP-SGD by better utilizing the available resource budget to increase the model accuracy. 

\subsection{Effectiveness of the Approximate Solution Approach for the DP-PASGD Optimal Design}
\begin{figure*}[ht]
\centering
\includegraphics[width=0.9\linewidth]{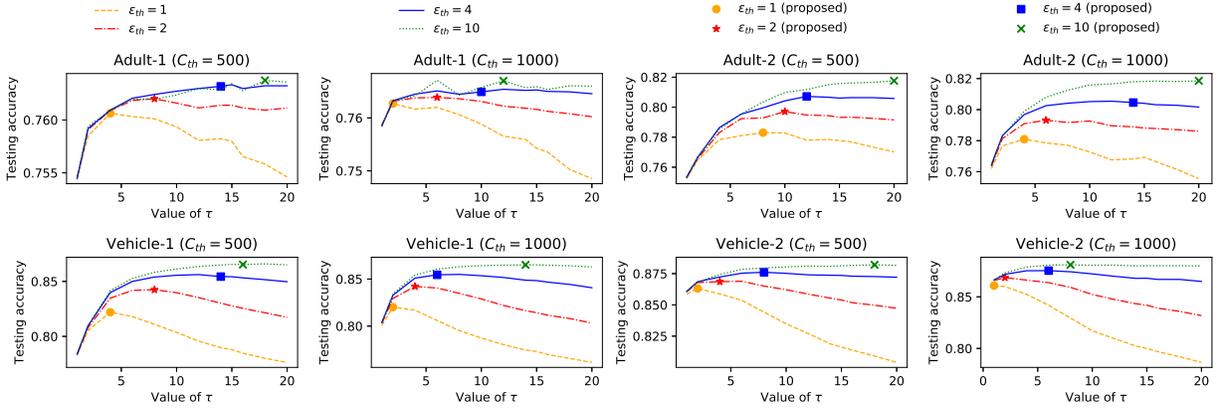}
\caption{Performance of DP-PASGD with different $\tau$ when resource budget $C_{th}=500$ or $ 1000$. The curves show the training loss and testing accuracy with different $\tau$. The single marker represents the result of our proposed approach with $\tau$ computed by the proposed optimization framework.}\label{fig:baseline_proposed}
\end{figure*}



In this subsection, we show the effectiveness of our approximate solution approach in finding the optimal configuration of DP-PASGD by comparing it with the configuration found by the brute-force method. Given a resource budget $C_{th}$ and a privacy budget $\epsilon_{th}$, we have to set the values of $K, \tau$, and $\sigma_m^2$ in order to start the training. It is common to use the grid search to tune these hyperparameters, which needs to try all combinations of these hyperparameters on the validation set and returns the combination with the highest testing accuracy as the optimal configuration. 
The grid search is a brute-force method which is costly, especially for sensitive datasets. In comparison, our approach can efficiently find the optimal configuration of DP-PASGD using the proposed optimization framework, saving both time and privacy costs in tuning the hyperparameters. 
Here, to demonstrate the effectiveness of our optimal design approach for DP-PASGD, we compare the optimal aggregation period $\tau$ calculated by our optimal design approach with the best $\tau$ obtained by using the grid search. 

Specifically, to do the grid search, we enumerate all possible values of $\tau$ ranging from 1 to 20 for each task. For each $\tau$, we tune $K$ from 200 to $ {C_{th}}/({c_1/\tau + c_2})$ on the validation set to find the optimal $K$ under the resource budget $C_{th}$. Note that once $K$ is determined, $\sigma_m^2$ can be determined by \eqref{eps_k}. Then, we show the best testing accuracy achieved by each $\tau$ under different resource and privacy budgets and then find the global optimal $\tau^*$. 

We compare our optimal design approach with the grid search method on all the data distribution cases under 2 resource budgets (i.e., $C_{th}=500$ or $1000$) and 4 privacy budgets (i.e., $ \epsilon_{th}=1,2,4$ or $10$).
The results are depicted in Figure~\ref{fig:baseline_proposed}. It is easy to see that there exists an optimal aggregation period $\tau^*$ that maximizes the learning performance under the privacy and resource budgets. We observe that the value of $\tau$ found by our proposed approach (represented by a single point in the figures) is very close to the optimal one obtained by the grid search under all cases, verifying the effectiveness of our approximate solution approach. 
%
%
Besides, we can observe that in all cases, when the privacy budget $\epsilon_{th}$ increases from 1 to 10, the optimal value of $\tau^*$ would almost always increase. On the other hand, when the resource budget $C_{th}$ changes from 500 to 1000, the optimal value of $\tau^*$ would almost always decrease. In Section~\ref{subsec:tau_rec_pr}, we show the values of optimal $\tau$ under different resource and privacy budgets.

\subsection{Trade-offs among Accuracy, Privacy and Cost}
\begin{figure}[ht]
\centering
\includegraphics[width=0.9\linewidth]{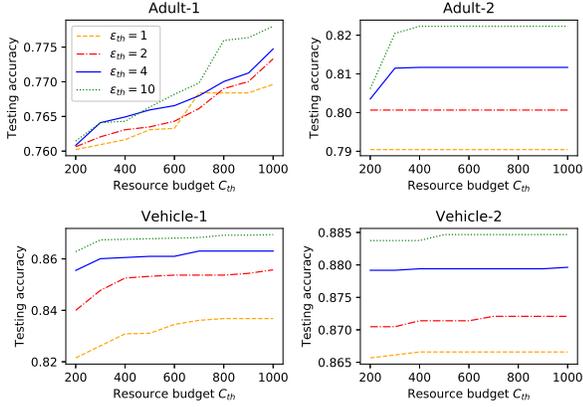}
\caption{Trade-off between resource budget and accuracy.}\label{fig:tradeoff_resource_acc}
\end{figure}
\begin{figure}[ht]
\centering
\includegraphics[width=0.9\linewidth]{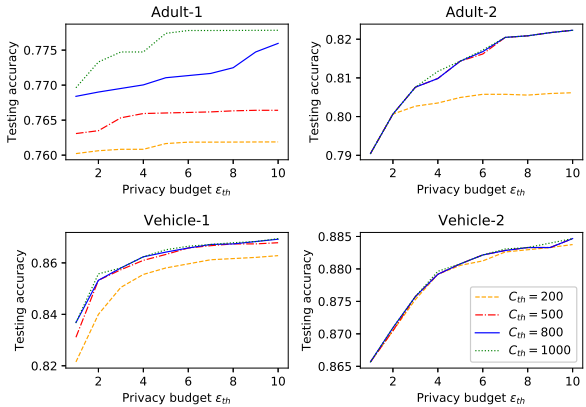}
\caption{Trade-off between privacy budget and accuracy.}\label{fig:tradeoff_privacy_acc}
\end{figure}


In this subsection, we evaluate the inherent trade-offs among model accuracy, resource cost, and privacy of federated learning under DP-PASGD. First, to show the trade-off between model accuracy and resource cost, we compute the testing accuracy of DP-PASGD achieved by our proposed optimal design under different resource budgets $C_{th}$ while fixing the privacy budget $\epsilon_{th}$. Specifically, we consider 4 different settings where $\epsilon_{th}=1,2,4$ or $10$ respectively. For each setting, we vary the the resource budget from 200 to 1000. The results are shown in Figure~\ref{fig:tradeoff_resource_acc}. From the figure, we can observe that higher resource budget generally implies higher testing accuracy. The reason is that when the resource budget $C_{th}$ is smaller, devices have less chance to improve their local models and reduce the divergence among their models via aggregating with others, and therefore the accuracy is lower. However, for the Adult-2, Vehicle-1, and Vehicle-2 cases, the testing accuracy does not significantly increase when more resources are allocated, especially when the privacy budget is low. The reason is that the models learned in those cases are more sensitive to the privacy budget. Therefore, even when we have more resources for computation and communication, the accuracy is still limited by the privacy budget.

Next, we show the trade-off between privacy and model accuracy by computing the testing accuracy with DP-PASGD under different privacy budgets $\epsilon_{th}$ while fixing the resource budget $C_{th}$. Here, we set $C_{th}=200,500,800$ or $1000$ for 4 different settings. For each setting, we vary the privacy budget from 1 to 10. The results are shown in Figure~\ref{fig:tradeoff_privacy_acc}. We can see that higher privacy budget usually leads to higher testing accuracy due to the decreased noise magnitude added in each iteration. However, for the Adult-1 case, the impact of the privacy budget is less significant cmpared to other data distributed cases because its model turns out to be more sensitive to the resource budget. 

\subsection{Impact of System Settings on $\tau$}\label{subsec:tau_rec_pr}

\begin{figure}[ht]
\centering
\includegraphics[width=0.9\linewidth]{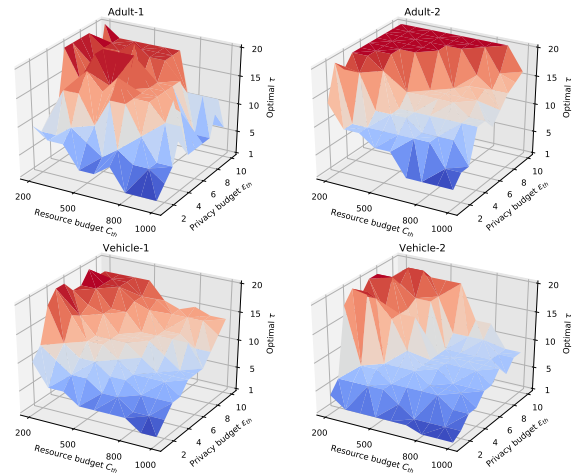}
\caption{The optimal global aggregation period $\tau$ with different resource and privacy budgets.}\label{fig:c_e_tau}
\end{figure}


In this subsection, we evaluate the impact of the setting of resource and privacy budgets on the optimal $\tau$. The change of the optimal $\tau$ in DP-PASGD with different resource budgets and privacy budgets is shown in Figure~\ref{fig:c_e_tau}. When the resource budget is large and privacy budget is small, DP-PASGD chooses a smaller $\tau$ to aggregate more frequently so as to reduce the iteration number, saving privacy loss. On the other hand, when the resource budget is small and privacy budget is large, DP-PASGD chooses a larger $\tau$ to save communication cost and do more local computation during each aggregation period. 


\section{Conclusion}\label{sec:con}
In this paper, we have proposed a novel privacy-preserving federated learning scheme, called DP-PASGD, for resource-constrained IoT. We have performed the convergence analysis of the proposed DP-PASGD and investigated the optimal configuration of DP-PASGD to maximize the model accuracy under resource and privacy limits. Extensive experiments based on real-world datasets have verified the effectiveness of the proposed scheme and shown the trade-offs among model accuracy, resource cost, and privacy for federated learning in IoT. In future work, we plan to study the performance of DP-PASGD in other learning settings such as multi-task learning and privacy considerations such as personalized differential privacy. 


\ifCLASSOPTIONcaptionsoff
  \newpage
\fi

\bibliographystyle{IEEEtran}
\bibliography{guo,gong,reference}

\end{document}